\pdfoutput=1

\documentclass[11pt]{article}

\usepackage[final]{acl}

\usepackage{times}
\usepackage{latexsym}

\usepackage[T1]{fontenc}

\usepackage[utf8]{inputenc}

\usepackage{microtype}

\usepackage{inconsolata}

\usepackage{times}
\usepackage{latexsym}
\usepackage{amsfonts}
\usepackage{amsmath}
\usepackage{subcaption}
\usepackage{graphicx}

\usepackage{xcolor}
\usepackage{CJKutf8}

\usepackage{tabularx}
\usepackage{booktabs}
\usepackage{multirow}

\usepackage{enumitem}

\title{Chain of Condition: Construct, Verify and Solve Conditions for Conditional Question Answering}

\author{
    Jiuheng Lin$^{1}$,
    Yuxuan Lai$^{2,3}$,
    Yansong Feng$^{1}$\thanks{Corresponding author.} \\
    $^{1}$ Wangxuan Institute of Computer Technology, Peking University, China\\
    State Key Laboratory of General Artificial Intelligence,  Peking University, China\\
    $^{2}$ Department of Computer Science, The Open University of China\\
    $^{3}$ Engineering Research Center of Integration and Application of Digital \\ Learning Technology, Ministry of Education\\
    {\tt linjiuheng@stu.pku.edu.cn} ~~
    {\tt erutan@pku.org.cn} ~~
    {\tt fengyansong@pku.edu.cn} \\
}

\begin{document}
\maketitle
\begin{abstract}
Conditional question answering (CQA) is an important task that aims to find probable answers and identify missing conditions. Existing approaches struggle with CQA due to two challenges: (1) precisely identifying necessary conditions and the logical relationship, and (2) verifying conditions to detect any that are missing. In this paper, we propose a novel prompting approach, Chain of condition, by first identifying all conditions and constructing their logical relationships explicitly according to the document, then verifying whether these conditions are satisfied, finally solving the logical expression to indicate any missing conditions and generating the answer accordingly. Experiments on two CQA benchmark datasets show our chain of condition outperforms existing prompting baselines, establishing a new state of the art. Furthermore, with only a few examples, our method can facilitate GPT-3.5-Turbo or GPT-4 to outperform all existing supervised models.\footnote{Our code is open sourced at: \url{https://github.com/Infinite-set/Chain-of-Condition}}

\end{abstract}

\section{Introduction}

\begin{figure}[t]
\centering
\includegraphics[scale=0.48]{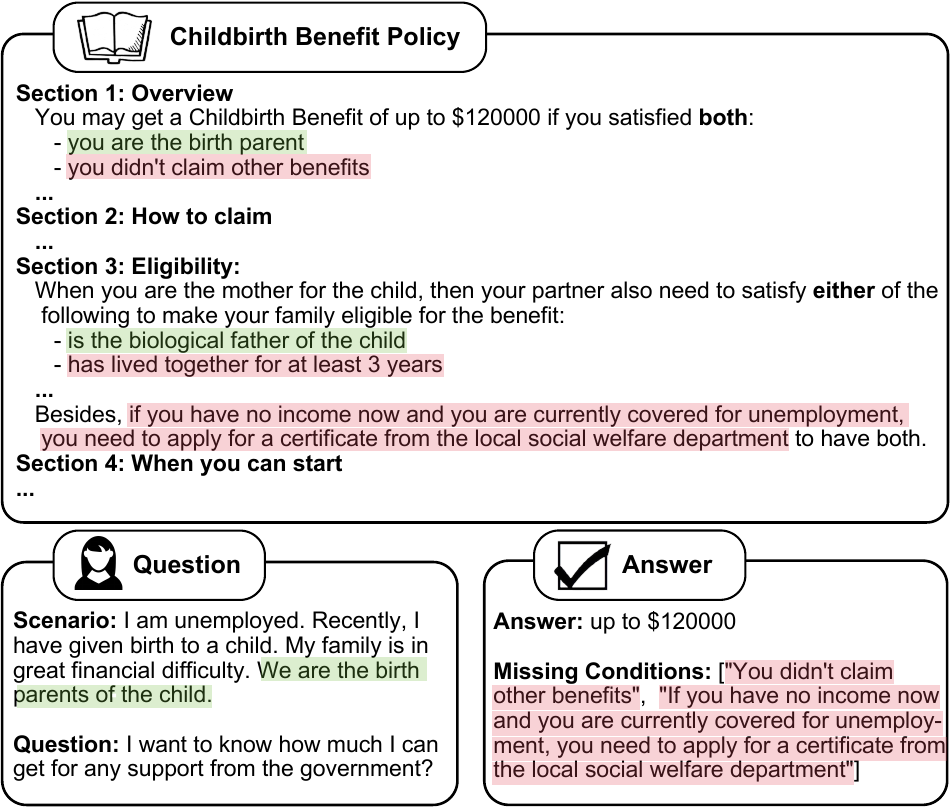}
\caption{An example of conditional question answering. All conditions are colored. The conditions in green are satisfied by the user's description, while those in red are not mentioned. The second red condition, \textit{has lived together for at least 3 years}, is not necessary because it has an "either" relationship with an already satisfied condition. But the other two red conditions are required to make the answer "up to \$120000" correct.}
\label{fig:cqa_example}
\end{figure}

Conditional question answering (CQA) aims to answer questions where the information provided by the user may not be sufficient, and any missing information should be requested from the user to determine the answer~\cite{saeidiInterpretationNaturalLanguage2018,min-etal-2020-ambigqa,sunConditionalQAComplexReading2021,Dhingra_2022,ju-etal-2022-cmqa,zhang-etal-2023-many}. CQA is a challenging and promising task, which has been gaining increasing attention recently~\cite{sun2022reasoning,du-etal-2023-structure,wangLearningStructuredDocuments2023,hussainLeveragingLLMsConditional2023,puertoCodePromptingElicits2024}.
Figure~\ref{fig:cqa_example} shows an example. The user asks for the amount of benefit she would receive, but according to the policy, the applicant must not claim for other benefits and has an unemployment certificate as prerequisite. 
These conditions are not mentioned in the user's description, therefore it is necessary to remind the user to provide this missing information to determine the benefit amount, which is \textit{up to \$120000}.

The mains challenges for CQA are twofold. The first one is identifying all conditions from the document and analyzing their relationship.  In Figure~\ref{fig:cqa_example}, all conditions are highlighted, while the majority of the description focuses on irrelevant aspects of Childbirth Benefit, making it difficult to accurately identify all relevant conditions. Moreover, there are specific relationships between conditions. For example, the conditions \textit{you are the birth parent} and \textit{you didn't claim other benefits} must  be satisfied \textbf{simultaneously}, whereas the conditions \textit{is the biological father of the child} and \textit{has lived together for at least 3 years} require \textbf{at least one} to be satisfied. Precisely identifying all conditions and their relationships according to the document is a prerequisite for the CQA task, but existing approaches primarily build end-to-end systems that overlook this challenge~\cite{ainslieETCEncodingLong2020a, izacardFiDLeveragingPassage2021,hussainLeveragingLLMsConditional2023}. These methods take the whole document as input, train models to implicitly identify conditions and parse their relationships, and directly output the answer along with any missing conditions. Consequently, due to limited reasoning abilities in existing models, these approaches struggle with questions involving multiple conditions and complex relationships among them. Besides, their solution path is implicit thus impossible for users to interpret.

Verifying conditions and solving their logical relationships is the second challenge~\cite{sun2022reasoning}. Each condition may be satisfied, contradicted or ignored by the user, and must be correctly verified. Furthermore, multiple conditions along with their logical relationship will form a logical expression, and solving this expression is necessary for accurately identifying all the conditions missing from the user's input. For example, in Figure~\ref{fig:cqa_example}, the conditions in green are satisfied according to the query, while those in red are not mentioned. Moreover, although the condition \textit{has lived together for at least 3 years} is not mentioned by the user, it has an "either" relationship with an already satisfied condition, \textit{is the biological father of the child}. Therefore, the condition \textit{has lived together for at least 3 years} is not necessary for user to satisfy. Determining the un-necessity of such condition requires correctly solving the logical expression of conditions. 
Previous works train models to verify and solve conditions simultaneously, requiring models to implicitly resolve the expression~\cite{du-etal-2023-structure,wangLearningStructuredDocuments2023} and leading to computational errors and reduced precision.

To address these challenges, we introduce \textbf{Chain of Condition}, a novel prompting framework for constructing, verifying and solving conditions in CQA tasks. Chain of condition include three main steps: first explicitly identifying all conditions and \textbf{constructing} the logical expression about the conditions according to the document,
next \textbf{verifying} whether each condition has been satisfied by the user, finally \textbf{solving} the logical expression precisely through tool use to indicate missing conditions and generating the appropriate answer.

We conduct experiments on two CQA benchmark datasets ConditionalQA~\cite{sunConditionalQAComplexReading2021} and ShARC~\cite{saeidiInterpretationNaturalLanguage2018}. The results show that chain of condition significantly outperforms all prompting baselines. And with backbone models like GPT-3.5-Turbo or GPT-4, our chain of condition even performs better than all supervised baselines.

Our contributions are summarized as follows:\vspace{1mm}\\
(1) We propose Chain of condition, an construct-verify-solve three-step prompting framework for conditional question answering. Experiments on benchmark datasets show our method outperforms existing prompting baselines, establishing a new state-of-the-art. \vspace{1mm}\\ 
(2) Our proposed framework can enhance backbone models like GPT-3.5-Turbo or GPT-4 to surpasses all fully supervised baselines, with only a few examples.\vspace{1mm}\\
(3) Our Chain of condition can create coherent and interpretable reasoning paths that are easier for humans to understand.

\begin{figure*}[t]
  \centering
  \includegraphics[height=9cm]{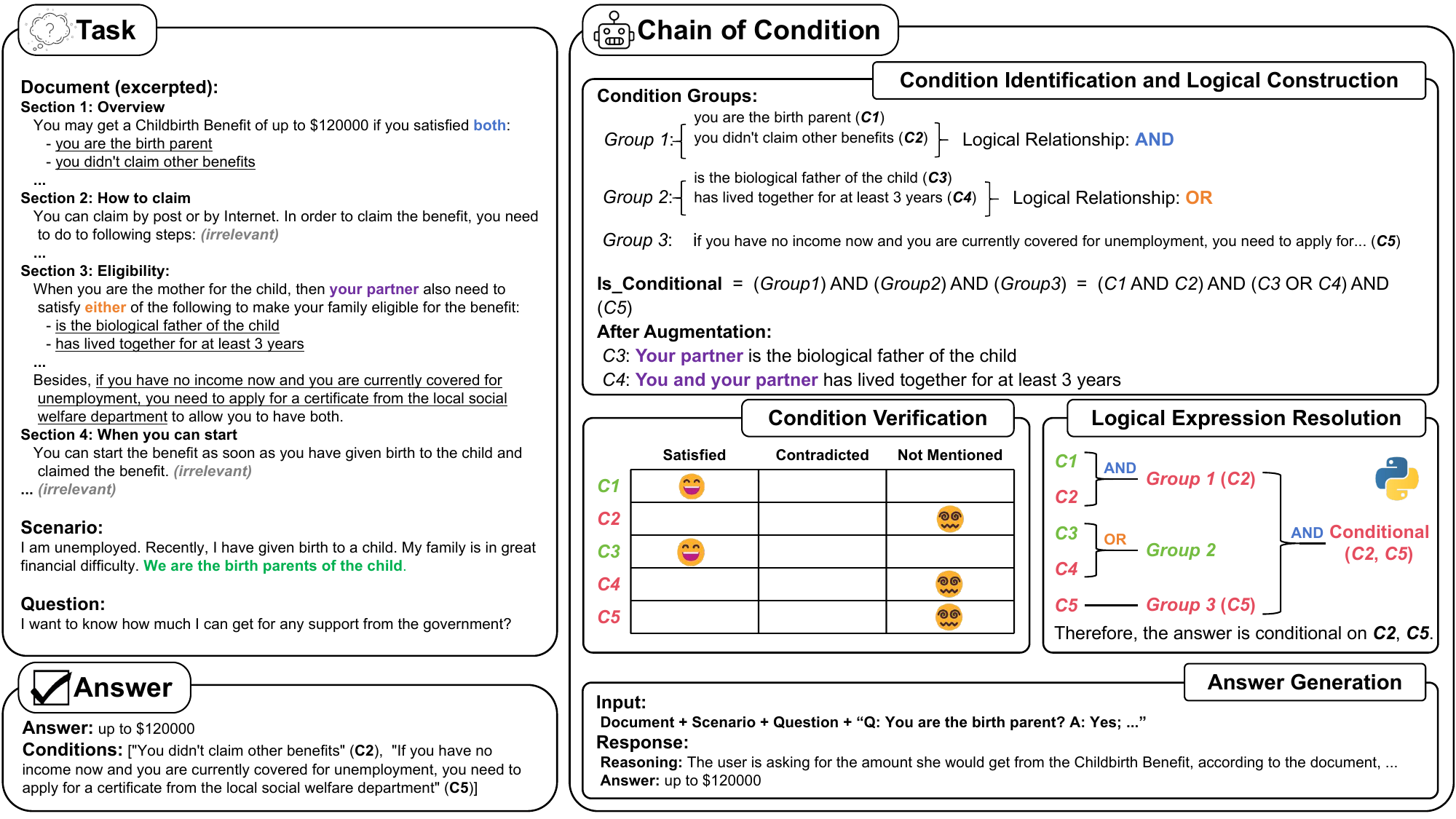}
  \caption{Method overview. Chain of condition consists of three main steps: condition identification and logical construction, condition verification, expression solution and answer generation.}
  \label{fig:method}
\end{figure*}

\section{Related Works} 
\vspace{1mm}
\noindent \textbf{Prompting Methods for LLMs    } Large language models can be guided to solve tasks in a step-by-step manner~\cite{wei2023chainofthought}. For more complex reasoning tasks such as multi-hop QA~\cite{yang2018hotpotqa} or math problems~\cite{cobbe2021training}, previous works typically address them by decomposing the question into simple sub-questions for models to solve sequentially~\cite{yaoReActSynergizingReasoning2023a,press-etal-2023-measuring,yu2023chainofnoteenhancingrobustnessretrievalaugmented,servantez2024chain,tao2024chainofdiscussion}. This decomposition reduces task difficulty and improves interpretability. Another approach to enhance performance on reasoning tasks is by combining LLMs with symbolic interpreters such as a Python runtime~\cite{chenProgramThoughtsPrompting2022,gao2023pal,lyu-etal-2023-faithful} or a SAT solver~\cite{ye2023satlm}. Compared to textual reasoning, program-based reasoning is executed accurately, thus achieving high-precision reasoning in complex questions. Furthermore,~\citet{puertoCodePromptingElicits2024} compared the performance of these two approaches and showed that introducing code in the prompt may elicit the reasoning ability for the CQA task.
In this work, we integrate the ideas of decomposing task and leveraging code interpreter into our framework, enabling it to benefit from both approaches' advantages. This not only improves interpretability but also increases precision.

\vspace{1mm}
\noindent \textbf{Supervised Methods for the CQA task   } Various pre-trained models have been proposed for the conditional question answering (CQA) task, including both extractive and generative models. Extractive models such as the ETC pipeline~\cite{ainslieETCEncodingLong2020a} and DocHopper~\cite{sunIterativeHierarchicalAttention2021} extract answers and conditions from input content. Generative models, including FiD~\cite{izacardFiDLeveragingPassage2021}, SDHG~\cite{du-etal-2023-structure}, and TReasoner~\cite{sun2022reasoning}, leverage generative models to directly generate answers and conditions together. Additionally, \citet{hussainLeveragingLLMsConditional2023} explored directly fine-tuning LLMs for the CQA task, demonstrating better performance but at a much higher training cost, while \citet{wangLearningStructuredDocuments2023} proposed the LSD framework to generate more conditional questions for fine-tuning. However, these methods are often limited to specific downstream fine-tuning tasks and lack generalizability. In contrast, chain of condition does not require further fine-tuning and exhibits better generalizability due to its few-shot setting.

\section{Preliminary}
We investigate the conditional question answering (CQA) task, where all missing condition should be requested from the user to determine the answer. Formally, the task's input consists of the user's \textbf{question} $Q$ and \textbf{scenario} $S$ providing some background information, and a reference \textbf{document} $D$ describing the policy being questioned. The answer should be inferred from the document. Unlike other QA tasks, the document in the CQA task contains many \textbf{conditions} $\mathbb{C} = \{c_1;c_2;...;c_n\} \subset D$ that must be satisfied by the user. The complete output includes the answer along with any unmentioned conditions $A = (a, \mathbb{C}^{\mathrm{(u)}})$, where $\mathbb{C}^{\mathrm{(u)}} = \{c_1^{\mathrm{(u)}};...;c_k^{\mathrm{(u)}}\} \subset \mathbb{C}$ denotes the $i$-th unmentioned condition for answer $a$, and $k_i \ge 0$ denotes the total number of unmentioned conditions for it\footnote{A few questions in ConditionalQA have multiple answers with conditions, and we leave the condition prediction for them as future work.}. If there are no unmentioned conditions, then we categorize the answer as \textit{deterministic}. Otherwise, we call it \textit{conditional}, and all missing conditions should be listed simultaneously with the answer.

\section{Methodology}
\label{sec:methodology}

We introduce \textbf{Chain of Condition}, a novel approach to guide Large Language Models (LLMs) for conditional question answering tasks. Our chain of condition pipeline is similar to the human process for dealing with such problems, thus can demonstrate a more coherent and interpretable solving path for users to read, as illustrated in Figure~\ref{fig:method}.

Chain of condition includes three steps: \textbf{condition identification and logical construction}, \textbf{condition verification}, \textbf{expression resolution and answer generation}.
Our framework decompose the original CQA task into smaller sub-tasks, allowing the model to solve them sequentially. First, we identify all conditions and recognize their logical relationships according to the document, forming a logical expression of conditions. Secondly, we verify each condition's fulfillment in the user's description. Finally, We take the verification results to solve the logical expression and identify all missing conditions through Python interpreter, and generate the answer by taking the condition solutions into account. A prompt example for each step is shown in Appendix~\ref{app:prompt}.

\subsection{Condition Identification and Logical Relationship Construction}

The document $D$ contains a substantial amount of irrelevant information, so the first step is to filter this out and identify all relevant conditions $\mathbb{C} = \{c_1;c_2;...;c_n\} \subset D$. In addition to identifying the conditions, it is crucial to arrange them in a particular structure according to the document. 
We address this by introducing condition groups $\mathbb{G} = \{G_1;G_2;...;G_k\}$, where $k$ represents the total number of condition groups in the document, and each group $G_i$ consists a set of nearby conditions sharing the same relationship. The $i$-th condition group $G_i$ is defined as $G_i(c_1^i,c_2^i,...,c_{n_i}^i,o_i) = (f(c_1^i)\ o_i\ f(c_2^i)\ o_i\ ...\ o_i\ f(c_{n_i}^i))$, where $\{c_1^i,...,c_{n_i}^i\} $ are the conditions in the $i$-th group, $o_i \in \{\mathrm{AND}, \mathrm{OR}\}$ is the logical operator connecting them and $f(c)$ represents the fulfillment of condition $c\in \mathbb{C}$ that will be verified in the next step. we instruct the model to directly identify the condition locations and their relationship in the document, then parse the model's output to obtain $\mathbb{G}$, ultimately forming a compositional logical expression of conditions $F(G_1,...,G_k,o) = ((G_1)\ o\ ...\ o\ (G_k))$, where $o \in \{\mathrm{AND}, \mathrm{OR}\}$. The solving result of $G_i$ and $F$ are in $\{\Bar{\textbf{d}},\Bar{\textbf{c}}\}$, where $\Bar{\textbf{d}}$ denote the \textit{deterministic} answer and $\Bar{\textbf{c}}$ \textit{conditional} answer. Detailed discussion of the solving process is in Section~\ref{subsec:exp_sol}

Besides, the conditions in the document are usually free-form, making it difficult to thoroughly separate a condition from other irrelevant context. Simply truncating the document may result in incomplete conditions. For example, in Figure~\ref{fig:method}, the condition \textit{is the biological father of the child} lacks a subject \textit{your partner}, which needs to be extracted from the previous sentence. This incompleteness could prevent the model from correctly understanding the meaning of the condition, consequently hindering its ability to accurately verify the fulfillment of the condition.

To address this problem, we take context-based augmentation after identifying conditions' locations. This approach allows us to obtain a short paragraph for each condition, containing all the necessary information. Specifically, when the document has a certain structure (e.g. subsections), we use this to find the relevant context for augmentation. We take the entire subsection where the condition appears as the augmentation paragraph, ensuring it contains enough background information while being much shorter than the entire document. When there is no structural information available, we instruct the model to directly summarize the condition based on the context.

\subsection{Condition Verification}

After acquiring all condition groups from $D$ and augmenting each condition, we instruct the models to sequentially verify the fulfillment of each condition based on $S$. This involves taking the question, scenario, and augmented condition as input, and leveraging LLMs to determine the fulfillment of conditions. For each condition $c \in \mathbb{C}$, the verification process can be formalized as determining the value of function $f(c) \in \{\Tilde{\textbf{s}}, \Tilde{\textbf{c}}, \Tilde{\textbf{n}}\}$. Here, $\Tilde{\textbf{s}}$ means the condition is satisfied by the user, $\Tilde{\textbf{c}}$ means it is contradicted, and $\Tilde{\textbf{n}}$ means the condition is not mentioned. Conditions that are either satisfied or contradicted lead to a \textit{deterministic} answer and can be treated as true values, while unmentioned conditions result in a \textit{conditional} answer and are treated as false. Therefore, solving the expression $F$ can be viewed as boolean operations on true/false values with special shortcut judgement. This process will be discussed further in Section~\ref{subsec:exp_sol}.

\subsection{Expression Solution and Answer Generation}
\label{subsec:exp_sol}

After obtaining the verification result $f(c)$ for each condition $c$, we take these results into groups $\mathbb{G}$ and logical expression $F$ to determine whether the answer is \textit{conditional} or \textit{deterministic}. And if the answer is \textit{conditional}, all missing conditions should be listed along with the answer.

Traditionally, this is done by prompting models to implicitly reason and resolve the logical expression. However, recent studies have shown that even large language models struggle with logical or mathematical reasoning tasks~\cite{blairstanek2023gpt3}. Therefore, a better solution is to offload the computation process to an external symbolic interpreter~\cite{chenProgramThoughtsPrompting2022,gao2023pal,lyu-etal-2023-faithful,ye2023satlm}. 

In chain of condition, we use a Python interpreter to solve the logical expression $F(G_1,...,G_k,o)$. Specifically, we treat all "not mentioned" conditions $\{c|f(c) = \Tilde{\textbf{n}}\}$ as false values, while "satisfied" or "contradicted" conditions $\{c|f(c) \in \{\Tilde{\textbf{s}}, \Tilde{\textbf{c}}\}\}$ are considered true. Additionally, a "satisfied" condition linked by an "OR" relationship in its condition group or a "contradicted" condition with an "AND" relationship in its condition group $G_i$ would form a shortcut, allowing for direct verification of $G_i$.
This process can be formally expressed as $\mathbb{G}^{\mathrm{sc}} = \left\{ G_i \mid o_i = \mathrm{or}, \, \exists c \in G_i \, \hspace{0.3em} \text{s.t.} \, f(c) = \Tilde{\textbf{s}} \right\} \cup \left\{ G_j \mid o_j = \mathrm{and}, \, \exists c \in G_j \, \hspace{0.3em} \text{s.t.} \, f(c) = \Tilde{\textbf{c}} \right\}$. Finally, a group $G_i$ is considered \textit{conditional} (i.e., $G_i = \Bar{\textbf{c}} $) if there exists unmentioned condition within it and no shortcut has been formed, formally $ G_i \notin \mathbb{G}^{\mathrm{sc}},\ \exists c \in G_i \, \hspace{0.3em} \text{s.t.} \, f(c) = \Tilde{\textbf{c}} $, and otherwise, it is classified as \textit{deterministic}. This reasoning process is automatically executed via Python using Boolean operations and special shortcut judgments.

Furthermore, we indicate all missing conditions, i.e., when a condition is not mentioned and all other conditions in the same group have not triggered any shortcut, formally $\mathbb{C}^{\mathrm{(u)}} = \{c_i^j|f(c_i^j)=\Tilde{\textbf{n}}, G_j=\Bar{\textbf{c}}, G_j \notin \mathbb{G}^{\mathrm{sc}},\forall i\le n_j,\forall j\le k\}$. If $\mathbb{C}^{\mathrm{(u)}} \ne \emptyset$, then the answer is \textit{conditional}, and otherwise \textit{deterministic}. By levaraging Python interpreter, we reduce model inference costs, improve precision, and enhance interpretability.

After obtaining the verification result for the conditions, we instruct the model to generate the answer. Since we have already verified each condition's fulfillment, we can leverage this information for more accurate answer generation. Specifically, we add these conditions $c$ along with their fulfillment $f(c)$ into the prompt. This provides the model with straightforward information about the conditions, reducing the need to infer their fulfillment from the document.

\begin{table*}[t]
\centering
\begin{small}
\setlength\tabcolsep{2pt}
\begin{tabular}{l|cc|cc|cc|cc|cc}
\toprule
\multirow{2}{*}{\textbf{Method}} & \multicolumn{2}{c|}{\textbf{GPT-3.5}} & \multicolumn{2}{c|}{\textbf{Llama-2 (70B)}} & \multicolumn{2}{c|}{\textbf{Llama-2 (13B)}} & \multicolumn{2}{c|}{\textbf{Mistral}} & \multicolumn{2}{c}{\textbf{Average}} \\
 & \textbf{EM/F1} & \textbf{w/conds} & \textbf{EM/F1} & \textbf{w/conds} & \textbf{EM/F1} & \textbf{w/conds} & \textbf{EM/F1} & \textbf{w/conds} & \textbf{EM/F1} & \textbf{w/conds} \\
\midrule
Zero-Shot & 59.5/\underline{71.0} & 23.9/29.5 & 44.0/51.2 & 26.6/30.9 & 42.3/49.6 & 26.1/28.9 & 44.2/50.7 & 26.7/30.8 & 47.5/55.6 & 25.8/30.0 \\
Chain of Thought & 59.3/70.0 & 45.4/54.6 & \underline{62.2}/\underline{71.4} & \underline{45.5}/\underline{53.7} & \underline{56.8}/\underline{65.8} & \underline{38.7}/\underline{44.8} & \textbf{58.3}/\textbf{68.6} & 37.7/46.4 & \underline{59.2}/\underline{69.0} & \underline{41.8}/\underline{49.9} \\
Code Prompting & \underline{60.4}/68.2 & \underline{50.8}/\underline{57.5} & 54.4/63.1 & 15.9/19.2 & 45.9/49.7 & 11.0/12.3 & 48.4/52.3 & 10.4/10.6 & 52.3/58.3 & 22.0/24.9  \\
Self-Ask & 54.9/66.9 & 41.3/52.2 & 59.2/69.9 & 36.1/45.5 & 47.9/59.9 & 30.3/38.3 & 49.6/60.5 & \textbf{41.2}/\textbf{50.4} & 52.9/64.3 & 37.2/46.6 \\
Chain of Condition & \textbf{64.6}/\textbf{73.7} & \textbf{52.9}/\textbf{61.0} & \textbf{64.7}/\textbf{75.2} & \textbf{47.7}/\textbf{56.0} & \textbf{57.2}/\textbf{67.1} & \textbf{43.0}/\textbf{51.3} & \underline{55.5}/\underline{63.8} & \underline{40.7}/\underline{47.5} & \textbf{60.5}/\textbf{70.0} & \textbf{46.1}/\textbf{54.0} \\
\bottomrule
\end{tabular}
\end{small}
\caption{Result of prompting methods on ConditionalQA. The best scores are made \textbf{bold}, with the second \underline{underlined}.}
\label{tab:llm_result_conditionalqa}
\end{table*}

\begin{table*}[t]
\centering
\begin{small}
\begin{tabular}{l|c|c|c|c|c}
\toprule
\makebox[0.15\textwidth][l]{\textbf{Method}} & \makebox[0.13\textwidth][c]{\textbf{GPT-3.5}} & \makebox[0.13\textwidth][c]{\textbf{Llama-2 (70B)}} & \makebox[0.13\textwidth][c]{\textbf{Llama-2 (13B)}} & \makebox[0.13\textwidth][c]{\textbf{Mistral}} & \makebox[0.13\textwidth][c]{\textbf{Average}} \\
\midrule
Zero-Shot & 63.2 & 43.8 & 45.5 & 36.9 & 47.4 \\
Chain of Thought & 66.7 & \underline{69.6} & 63.0 & 60.2 & 64.9 \\
Code Prompting & 60.4 & 39.9 & 37.6 & 40.3 & 44.6 \\
Self-Ask & \textbf{70.3} & 69.1 & \textbf{67.4} & \underline{60.5} & \underline{66.8} \\
Chain of Condition & \underline{70.2} & \textbf{74.9} & \underline{64.2} & \textbf{61.8} & \textbf{67.8} \\
\bottomrule
\end{tabular}
\end{small}
\caption{Result of prompting methods on ShARC. The best scores are made \textbf{bold}, with the second \underline{underlined}.}
\label{tab:llm_result_sharc}
\end{table*}

\begin{table}[t]
\centering
\begin{small}
\begin{tabular}{l|cc}
\toprule
\textbf{Method} & \textbf{EM/F1} & \textbf{w/conds} \\
\midrule
\multicolumn{3}{l}{Supervised Baselines} \\
\midrule
SDHG & 49.0/56.5 & 39.0/46.0 \\
TReasoner & 57.2/63.5 & \underline{46.1}/\underline{51.9} \\
LSD+Longformer & \underline{58.7}/\underline{66.2} & 45.0/50.5 \\
\midrule
\multicolumn{3}{l}{Chain of Condition} \\
\midrule
GPT-3.5 (Retrieval) & 56.6/\underline{66.2} & 42.1/51.0 \\
GPT-3.5 (16K) & \textbf{61.0}/\textbf{70.0} & \textbf{48.5}/\textbf{56.0} \\
\midrule[0.1pt]
GPT-3.5 (Oracle) & 64.6/73.7 & 52.9/61.0 \\
GPT-4 (Oracle) & 70.8/79.5 & 56.9/63.0 \\
\bottomrule
\end{tabular}
\end{small}
\caption{Chain of condition vs. supervised baselines.}
\label{tab:supervised_result}
\end{table}

\section{Experimental Setup}

\subsection{Datasets and Evaluation Metrics}
\label{subsec:exp:datasets}

Throughout our experiments, we use two conditional question answering datasets: ConditionalQA~\cite{sunConditionalQAComplexReading2021} and ShARC~\cite{saeidiInterpretationNaturalLanguage2018}. More dataset information is in Appendix~\ref{app:dataset}.

\noindent \textbf{ConditionalQA} features long and complex documents, and has many different types of questions. Documents in ConditionalQA are usually well-structured with HTML tags for each paragraph, because they are grouped into sections and subsections describing public policies in the UK. This brings the convenience for condition identification and augmentation. 

Following previous works in CQA~\cite{sunConditionalQAComplexReading2021},  we use the original sets of metrics: EM/F1 and conditional EM/F1 (abbreviated as \textbf{w/conds}). EM measures the exact match of predicted answer spans with gold ones, while F1 is the harmonic mean of token-level precision and recall. Conditional EM/F1 jointly measures the correctness of answer spans and the predicted conditions, providing a more comprehensive assessment of a model's performance on the CQA task. The calculation formulas are in Appendix~\ref{app:metric}.

\noindent \textbf{ShARC} is a conversational QA dataset, and the original task is to answer the question if the information in the dialog history is enough, or to generate a new question to acquire missing information. We follow the previous work~\cite{puertoCodePromptingElicits2024} to isolate the QA task from the conversational setting to form a benchmark of the CQA task, resulting in a dataset that the model only needs to answer "yes", "no" or "not enough information". Additionally, we discard all irrelevant questions from the dataset for better evaluation. 

We use the accuracy on the answer prediction as the metric. Since there are no human annotated conditions in ShARC, so it is not possible to further measure the accuracy of predicted missing conditions. We leave more accurate evaluation for ShARC as future work.

\subsection{Baselines}

\paragraph{Prompting Baselines}
We compare our approach with 4 different prompting baselines in total. 

$\bullet$ \textbf{Code Prompting}~\cite{puertoCodePromptingElicits2024} is the only available prompting approach for the CQA task as far as we know. This method extends the original text prompt with additional LLM-generated code, which elicits the model's conditional reasoning abilities for CQA tasks.

$\bullet$ \textbf{Self-Ask}~\cite{press-etal-2023-measuring} is a recently proposed, well-performing prompting method, which we adapt for our CQA scenario. This method decomposes the question by explicitly asking and answering intermediate questions until reaching the final answer and missing conditions.

Additionally, we use \textbf{Zero Shot} and \textbf{Chain of Thought} prompting~\cite{wei2023chainofthought}  as baselines.
\paragraph{Supervised Baselines}
We include \textbf{SDHG}~\cite{du-etal-2023-structure}, \textbf{TReasoner}~\cite{sun2022reasoning} and \textbf{LSD}~\cite{wangLearningStructuredDocuments2023} as baselines for ConditionalQA.

For ShARC, since we follow previous work to modify the dataset's output format and discard all irrelevant instances\cite{puertoCodePromptingElicits2024}, there are no supervised baselines available. Therefore, we only compare chain of condition with other prompting baselines mentioned above.

\subsection{LLM Setup}

We conduct our experiments on four different LLMs to investigate whether our chain of condition performs consistently better across various settings. We use a commercial model, GPT-3.5-Turbo, and three open-source models, Llama-2-70B-chat, Llama-2-13B-chat, and Mistral-7B. Additionally, we leverage GPT-4~\cite{openai2023gpt4} for limited experiments exclusively on ConditionalQA due to cost constraints. For all models, we set the temperature to 0.0 to ensure reproducibility of the results, while using default settings for others. More details are in Appendix~\ref{app:setup}.

The original documents in ConditionalQA can be up to 9,320 tokens long, exceeding the context limitations of many LLMs, posing a challenge for all prompting methods. This could be solved by introducing a retriever to retrieve only relevant paragraphs of the document. 
When comparing with different prompting approaches, we use an oracle retriever to select relevant passages for a given question, to eliminate the interference from the retriever performance.
Following previous work~\cite{puertoCodePromptingElicits2024}, we retain all sections that include at least one human-annotated gold evidence and concatenating them to form the input.

And for comparison with supervised methods, we employ two approaches: (1) Using a retriever to retrieve relevant paragraphs from the document, and (2) Using a long-context version of an LLM as our backbone model.

\section{Results and Analysis}

\subsection{Main Results}
\label{sec:main_results}

We report the performance of chain of condition and all baselines on two benchmark datasets\footnote{The test set of ConditionalQA is not publicly available, and the question number is larger than dev set, causing a much higher API cost. Thus we only evaluate all methods on dev set.}.
Table~\ref{tab:llm_result_conditionalqa} presents the performance of all prompting methods on ConditionalQA, while Table~\ref{tab:llm_result_sharc} shows the performance on ShARC. Table~\ref{tab:supervised_result} compares the results of chain of condition with all supervised baselines on ConditionalQA.

The original evaluation script of ConditionalQA provides a detailed breakdown by question type. We report the overall results here, with all details in Appendix~\ref{app:results}.

\vspace{1mm}
\noindent \textbf{Chain of condition outperforms all prompting baselines on each dataset.} It establishes a new state-of-the-art. 
Additionally, Self-Ask also performs relatively well on ShARC, which can be attributed to the dataset’s features. ShARC consists 
of conversations that repeatedly ask for the questioner's information, which is very similar to the format of Self-Ask and thus naturally suitable. Therefore, it is reasonable to use Self-Ask for more fluent reasoning, leading to better performance.

\vspace{1mm}
\noindent \textbf{Chain of condition outperforms all supervised baselines.} With backbone models like GPT-3.5-Turbo or GPT-4, our framework surpasses all supervised baselines with few-shot settings.
This result highlights the promising future of prompting methods for the CQA task, not only achieving better performance but also reducing the costs for fine-tuning.

\vspace{1mm}
\noindent \textbf{Chain of condition remains effective when the input exceeds the LLM's context limit.} The result in Table~\ref{tab:supervised_result} show that using a retriever that better explores the document structures can achieve performance comparable to supervised baselines, while long-context LLMs can surpass all supervised baselines, demonstrating the feasibility of both methods when the input goes longer. Additionally, the time and monetary costs of long-context LLMs are higher than leveraging a retriever. Given limited computational resources, we believe that developing better retrievers is a more promising direction for future research.

\subsection{Analysis}
\label{subsec:analysis}

In this section, we first conduct ablation studies with GPT-3.5 on the chain of condition framework to demonstrate the necessity of each step. Next, we show that chain of condition consistently outperforms all baselines in more challenging task settings, and finally analyze the reasons for its superior performance.

\paragraph{Explicitly identifying and constructing conditions is crucial.} We make two hypotheses about why this step is essential. The first hypothesis is that it helps the model identify all possible conditions and 
ensures explicitly solution for logical expressions by external tools, both improving performance on conditions. Secondly, it allows us to generate the answer with additional information about 
condition verification results, which is only feasible if all conditions are explicitly identified.

To prove the first hypothesis, we conduct an ablation study on ConditionalQA since it has gold-labeled missing conditions. In this study, we prompted GPT-3.5 to perform end-to-end generation, i.e., first identify all conditions, then check their fulfillment, and finally indicate all unmentioned conditions implicitly through reasoning. As shown in Table~\ref{tab:ablation_1_conditionalqa}, this ablation results in a drop of 3.2 EM score and 2.8 F1 score for answer correctness, as well as 17.4 EM score and 19.5 F1 score for joint answer and condition correctness. The performance drop is much greater when evaluating both answers and conditions compared to evaluating answers alone, which indicates that removing the condition identification step leads to a much larger decrease in accuracy for conditions. Further investigation into the model's output reveals that the average number of predicted missing conditions for \textit{conditional} answers increases from 1.27 to 1.67, suggesting that GPT-3.5 tends to treat a condition as not mentioned by the user more frequently when the condition identification step is omitted.

We also conduct an ablation for the necessity of using both logical operators "AND" and "OR". We remove each of them and prompt the model using chain of condition. The results in Table~\ref{tab:ablation_1_conditionalqa} indicate removing either operator reduces the performance.

The second hypothesis is discussed in the ablation of the answer generation step.

\begin{table}[!t]
\footnotesize
\centering
\setlength\tabcolsep{1.8pt}
\begin{tabular}{l|cccc}
\toprule
\makebox[0.01\textwidth][c]{} & \makebox[0.07\textwidth][c]{\textbf{EM}} & \makebox[0.07\textwidth][c]{\textbf{F1}} & \makebox[0.07\textwidth][c]{\textbf{Cond EM}} & \makebox[0.07\textwidth][c]{\textbf{Cond F1}} \\
\midrule
Chain of Condition & 64.6 & 73.7 & 52.9 & 61.0 \\
\midrule
Prompting Only & 61.4 & 70.9 & 35.9 & 42.9 \\
$\Delta$ & \textit{-3.2} & \textit{-2.8} & \textit{-17.0} & \textit{-18.1} \\
\midrule
AND Only & 62.2 & 71.0 & 48.9 & 56.3 \\
$\Delta$ & \textit{-2.4} & \textit{-2.7} & \textit{-4.0} & \textit{-4.7} \\
\midrule
OR Only & 61.2 & 69.9 & 40.1 & 45.6 \\
$\Delta$ & \textit{-3.4} & \textit{-3.8} & \textit{-12.8} & \textit{-15.4} \\
\bottomrule
\end{tabular}
\caption{Ablation study for explicitly identifying and constructing conditions on ConditionalQA.}
\label{tab:ablation_1_conditionalqa}
\end{table}

\paragraph{Condition augmentation improves verification accuracy.} The removal of contextual information can hinder the model's ability to correctly understand the meaning of a condition. To prove this, we remove all other paragraphs of the condition's subsection, keeping only the original condition as input for verification on ConditionalQA. The result of conditional EM drops by 2.4 from 52.9 to 50.5, and the conditional F1 drops by 2.5 from 61.0 to 58.5 for this setting on GPT-3.5, indicating that the performance of condition prediction decreases due to reduced verification accuracy.

\begin{table}[t]
\centering
\begin{small}
\setlength\tabcolsep{5pt}
\begin{tabular}{l|cc|c}
\toprule
\multirow{2}{*}{} & \multicolumn{2}{c|}{\textbf{ConditionalQA}} & \textbf{ShARC} \\
 & \textbf{EM/F1} & \textbf{w/conds} & \textbf{Accuracy} \\
\midrule
Chain of Condition & 64.6/73.7 & 52.9/61.0 & 70.2 \\
\midrule
w/o Results & 61.4/70.9 & 50.5/59.1 & 67.5 \\
$\Delta$ & \textit{-3.2/-2.8} & \textit{-2.4/-1.9} & \textit{-2.7} \\
\bottomrule
\end{tabular}
\end{small}
\caption{Ablation study for answer generation on ConditionalQA and ShARC. \textit{w/o Results} refers to removing condition verification results from the answer generation input.}
\label{tab:ablation_3_condition}
\end{table}

\begin{table}[!t]
\footnotesize
\centering
\setlength\tabcolsep{1.8pt}
\begin{tabular}{l|cccc}
\toprule
\makebox[0.01\textwidth][c]{} & \makebox[0.07\textwidth][c]{\textbf{EM}} & \makebox[0.07\textwidth][c]{\textbf{F1}} & \makebox[0.07\textwidth][c]{\textbf{Cond EM}} & \makebox[0.07\textwidth][c]{\textbf{Cond F1}} \\
\midrule
Chain of Condition & 64.6 & 73.7 & 52.9 & 61.0 \\
\midrule
w/o Interpreter & 64.6 & 73.7 & 36.7 & 43.4 \\
$\Delta$ & - & - & \textit{-16.2} & \textit{-17.6}  \\
\bottomrule
\end{tabular}
\caption{Ablation study for leveraging the Python interpreter on ConditionalQA.}
\label{tab:ablation_python_solver}
\end{table}

\paragraph{Including verification results helps answer generation.} 
\label{paragraph: condition-aid generation analysis}
In this ablation, we remove the verification results of conditions from the input of answer generation. The results are shown in Table~\ref{tab:ablation_3_condition}. The performance drops by 3.2 EM score and 2.8 F1 score for answers, and by 2.4 EM score and 1.9 F1 score when jointly evaluating answers and conditions on ConditionalQA. Additionally, the accuracy drops by 2.7 on ShARC.

Furthermore, we find that the performance drop on ConditionalQA is mostly attributed to the yes/no questions, with a drop of 7.0 EM/F1 score and 5.3 conditional EM/F1 score. The likely reason for this phenomenon lies in the answer determination procedure: a span-type answer can be extracted directly in the document even without verifying any conditions. However, a yes/no answer must be inferred from the document along with each condition's fulfillment. Therefore, including the conditions' fulfillment in the prompt helps the model by reducing the need to repeatedly infer their fulfillment, allowing it to directly synthesize the information to generate the final answer.

\paragraph{Leveraging logical expression interpreter increases accuracy for conditions.} To demonstrate the importance of utilizing external symbolic solver for logical expressions, we conduct an ablation study where the Python interpreters is replaced with GPT-3.5. Since the resolution of logical expressions does not affect the answer generation process, the EM and F1 scores for answers alone remain consistent. However, when jointly evaluating both answers and conditions, a performance drop  is observed in Table~\ref{tab:ablation_python_solver}.

\paragraph{Chain of condition better finds missing conditions.} Most questions in these CQA datasets involve identifying and solving conditions, but only a small portion of them are truly \textit{conditional}. This is because, in many cases, the conditions for the answer are all satisfied by the user's scenario, so the model only needs to give a correct judgement on whether the answer is \textit{conditional}. However, when we consider only the \textit{conditional} answers in the dataset, correctly addressing them becomes more challenging. This is because the model not only needs to properly generate the answer and determine it as   \textit{conditional}, but also precisely finds the missing conditions.

The performance of all methods on ConditionalQA\footnote{ShARC does not have human-annotated conditions, so we could not experiment on it.} greatly drops when considering only the \textit{conditional} answers, as shown in Table~\ref{tab:analysis_1_conditional}. But, our chain of condition consistently outperforms other prompting baselines in this setting, demonstrating its effectiveness in finding missing conditions. 

Furthermore, in order to analyze the reasons behind chain of condition's superior performance on the CQA task, we divide the dev set of ConditionalQA based on the total number of gold conditions for each question in the document, resulting in two question groups. The first group contains data with at most one conditions, while the second group has at least three conditions, indicating a more complex set of conditions for solving.

We report the performance of GPT-3.5 with all prompting methods on these two groups in Table~\ref{tab:analysis_2_conds_num}. Since there is not a metric that directly measures the quality of predicted conditions, we additionally report the F1 score of the predicted conditions. The results highlight the increased difficulty of questions involving complex conditions, and chain of condition shows much less performance degradation in this more challenging group. This indicates its superior ability to handle complex conditions. We attribute this to the explicit identification of conditions and the use of a code interpreter to resolve the logical relationships among conditions.

\begin{table}[t]
\footnotesize
\centering
\setlength\tabcolsep{3pt}
\begin{tabular}{l|cccc}
\toprule
\textit{\textbf{Conditional}} & \makebox[0.07\textwidth][c]{\textbf{EM}} & \makebox[0.07\textwidth][c]{\textbf{F1}} & \makebox[0.07\textwidth][c]{\textbf{Cond EM}} & \makebox[0.07\textwidth][c]{\textbf{Cond F1}} \\
\midrule
Zero-Shot & 40.7 & 49.1 & 12.9 & 16.0 \\
CoT & 45.8 & 53.6 & 13.1 & 16.4 \\
Code & 47.2 & 54.3 & 8.5 & 11.5 \\
Self-Ask & 49.7 & 58.3 & 13.5 & 17.5 \\
Ours & \textbf{56.0} & \textbf{62.2} & \textbf{18.9} & \textbf{20.7} \\
\bottomrule
\end{tabular}
\caption{Result of different prompting methods on \textit{conditional} answer questions.}
\label{tab:analysis_1_conditional}
\end{table}

\begin{table}[t]
\footnotesize
\centering
\setlength\tabcolsep{2pt}
\begin{tabular}{l|cc|cc}
\toprule
 \textbf{\#Conds} & \multicolumn{2}{c|}{\textbf{<=1}} & \multicolumn{2}{c}{\textbf{>=3}} \\
 \textbf{Groups} &  \makebox[0.09\textwidth][c]{\textbf{EM/F1}} &  \makebox[0.09\textwidth][c]{\textbf{C\_F1}} &  \makebox[0.09\textwidth][c]{\textbf{EM/F1}} &  \makebox[0.09\textwidth][c]{\textbf{C\_F1}} \\
\midrule
Zero-Shot & 64.0/76.1 & 53.8 & 47.7/50.7 & 8.9 \\
CoT & 60.5/71.6 & 77.3 & 44.0/47.7 & 16.5 \\
Code & 61.9/70.1 & \textbf{87.9} & 51.7/57.0 & 4.0 \\
Self-Ask & 55.0/68.1 & 77.6 & 54.0/59.1 & 20.1 \\
Ours & \textbf{65.5}/\textbf{75.3} & 84.5 & \textbf{60.0}/\textbf{65.2} & \textbf{31.6} \\
\bottomrule
\end{tabular}
\caption{Performance on 2 groups in ConditionalQA. \textit{C\_F1} is the F1 score of predicted conditions.}
\label{tab:analysis_2_conds_num}
\end{table}

\section{Conclusion}

In this work, we propose Chain of condition, a novel prompting approach for conditional question answering. Our approach prompts LLMs to identify conditions, organize them into logical expressions and introduces a Python interpreter for resolution, effectively improving precision and enhancing interpretability. Experiments on all benchmark datasets show that chain of condition outperforms existing prompting baselines. Additionally, our method surpasses supervised baselines when utilizing a strong backbone model, demonstrating the promising future of prompting LLMs for CQA and paving the way for future research directions.

\section*{Limitations}

\paragraph{Retriever Performance} Although our chain of condition outperforms all baselines, we find that the retriever's performance (with a recall of around 70\%) is still a limiting factor. In this paper, we simply take an commercial embedding model (text-embedding-ada-002) as our retriever, and when the document length exceeds the LLM's input context capacity, we find this off-the-shelf retriever still overlook some necessary conditions and evidence. We believe employing a stronger retriever would achieve better performance. 

\paragraph{Token Efficiency} The multi-step prompting framework in chain of condition decomposes the original CQA task into several sub-tasks, thus leading to lower token efficiency compared to simpler prompting baselines and typically requiring more tokens to solve the entire problem.

\paragraph{Exploring Intermediate Results} The generated verification results can be better exploited. For instance, if the answer type is yes/no, then it is possible to directly leverage the condition verification results to determine the answer. This approach would not only enhance accuracy but also improve token efficiency, making it a promising direction for future research. Additionally, these intermediate results could be used as training data to distill our chain of condition framework into smaller models.

\section*{Acknowledgments}
This work is supported in part by NSFC (62206070, 62161160339) and Beijing Science and Technology Program (Z231100007423011).
We thank Chen Zhang, Mingxu Tao, Zirui Wu and Xiao Liu for their help in this work. Also, we thank Yixun Zhou and Kangcheng Luo for valuable suggestions in writing. For any correspondence, please contact Yansong Feng.

\bibliography{anthology,custom}

\begin{thebibliography}{28}
\providecommand{\natexlab}[1]{#1}

\bibitem[{Ainslie et~al.(2020)Ainslie, Ontanon, Alberti, Cvicek, Fisher, Pham, Ravula, Sanghai, Wang, and Yang}]{ainslieETCEncodingLong2020a}
Joshua Ainslie, Santiago Ontanon, Chris Alberti, Vaclav Cvicek, Zachary Fisher, Philip Pham, Anirudh Ravula, Sumit Sanghai, Qifan Wang, and Li~Yang. 2020.
\newblock \href {https://doi.org/10.48550/arXiv.2004.08483} {{{ETC}}: {{Encoding Long}} and {{Structured Inputs}} in {{Transformers}}}.
\newblock \emph{Preprint}, arxiv:2004.08483.

\bibitem[{Blair-Stanek et~al.(2023)Blair-Stanek, Holzenberger, and Durme}]{blairstanek2023gpt3}
Andrew Blair-Stanek, Nils Holzenberger, and Benjamin~Van Durme. 2023.
\newblock \href {https://arxiv.org/abs/2302.06100} {Can gpt-3 perform statutory reasoning?}
\newblock \emph{Preprint}, arXiv:2302.06100.

\bibitem[{Chen et~al.(2023)Chen, Ma, Wang, and Cohen}]{chenProgramThoughtsPrompting2022}
Wenhu Chen, Xueguang Ma, Xinyi Wang, and William~W. Cohen. 2023.
\newblock \href {https://arxiv.org/abs/2211.12588} {Program of thoughts prompting: Disentangling computation from reasoning for numerical reasoning tasks}.
\newblock \emph{Preprint}, arXiv:2211.12588.

\bibitem[{Cobbe et~al.(2021)Cobbe, Kosaraju, Bavarian, Chen, Jun, Kaiser, Plappert, Tworek, Hilton, Nakano, Hesse, and Schulman}]{cobbe2021training}
Karl Cobbe, Vineet Kosaraju, Mohammad Bavarian, Mark Chen, Heewoo Jun, Lukasz Kaiser, Matthias Plappert, Jerry Tworek, Jacob Hilton, Reiichiro Nakano, Christopher Hesse, and John Schulman. 2021.
\newblock \href {https://arxiv.org/abs/2110.14168} {Training verifiers to solve math word problems}.
\newblock \emph{Preprint}, arXiv:2110.14168.

\bibitem[{Dhingra et~al.(2022)Dhingra, Cole, Eisenschlos, Gillick, Eisenstein, and Cohen}]{Dhingra_2022}
Bhuwan Dhingra, Jeremy~R. Cole, Julian~Martin Eisenschlos, Daniel Gillick, Jacob Eisenstein, and William~W. Cohen. 2022.
\newblock \href {https://doi.org/10.1162/tacl_a_00459} {Time-aware language models as temporal knowledge bases}.
\newblock \emph{Transactions of the Association for Computational Linguistics}, 10:257–273.

\bibitem[{Du et~al.(2023)Du, Feng, Li, Li, Lan, and Zhao}]{du-etal-2023-structure}
Haowei Du, Yansong Feng, Chen Li, Yang Li, Yunshi Lan, and Dongyan Zhao. 2023.
\newblock \href {https://doi.org/10.18653/v1/2023.findings-acl.391} {Structure-discourse hierarchical graph for conditional question answering on long documents}.
\newblock In \emph{Findings of the Association for Computational Linguistics: ACL 2023}, pages 6282--6293, Toronto, Canada. Association for Computational Linguistics.

\bibitem[{Gao et~al.(2023)Gao, Madaan, Zhou, Alon, Liu, Yang, Callan, and Neubig}]{gao2023pal}
Luyu Gao, Aman Madaan, Shuyan Zhou, Uri Alon, Pengfei Liu, Yiming Yang, Jamie Callan, and Graham Neubig. 2023.
\newblock \href {https://arxiv.org/abs/2211.10435} {Pal: Program-aided language models}.
\newblock \emph{Preprint}, arXiv:2211.10435.

\bibitem[{Hussain et~al.(2023)Hussain, Dakle, Rallabandi, and Raghavan}]{hussainLeveragingLLMsConditional2023}
Syed-Amad Hussain, Parag~Pravin Dakle, SaiKrishna Rallabandi, and Preethi Raghavan. 2023.
\newblock \href {https://doi.org/10.48550/arXiv.2312.01143} {Towards leveraging {{LLMs}} for {{Conditional QA}}}.
\newblock \emph{Preprint}, arxiv:2312.01143.

\bibitem[{Izacard and Grave(2021)}]{izacardFiDLeveragingPassage2021}
Gautier Izacard and Edouard Grave. 2021.
\newblock \href {https://doi.org/10.18653/v1/2021.eacl-main.74} {({{FiD}}){{Leveraging Passage Retrieval}} with {{Generative Models}} for {{Open Domain Question Answering}}}.
\newblock In \emph{Proceedings of the 16th {{Conference}} of the {{European Chapter}} of the {{Association}} for {{Computational Linguistics}}: {{Main Volume}}}, pages 874--880, Online. Association for Computational Linguistics.

\bibitem[{Ju et~al.(2022)Ju, Wang, Zhang, Zheng, Liu, and Zhao}]{ju-etal-2022-cmqa}
Yiming Ju, Weikang Wang, Yuanzhe Zhang, Suncong Zheng, Kang Liu, and Jun Zhao. 2022.
\newblock \href {https://aclanthology.org/2022.coling-1.146} {{CMQA}: A dataset of conditional question answering with multiple-span answers}.
\newblock In \emph{Proceedings of the 29th International Conference on Computational Linguistics}, pages 1697--1707, Gyeongju, Republic of Korea. International Committee on Computational Linguistics.

\bibitem[{Lyu et~al.(2023)Lyu, Havaldar, Stein, Zhang, Rao, Wong, Apidianaki, and Callison-Burch}]{lyu-etal-2023-faithful}
Qing Lyu, Shreya Havaldar, Adam Stein, Li~Zhang, Delip Rao, Eric Wong, Marianna Apidianaki, and Chris Callison-Burch. 2023.
\newblock \href {https://doi.org/10.18653/v1/2023.ijcnlp-main.20} {Faithful chain-of-thought reasoning}.
\newblock In \emph{Proceedings of the 13th International Joint Conference on Natural Language Processing and the 3rd Conference of the Asia-Pacific Chapter of the Association for Computational Linguistics (Volume 1: Long Papers)}, pages 305--329, Nusa Dua, Bali. Association for Computational Linguistics.

\bibitem[{Min et~al.(2020)Min, Michael, Hajishirzi, and Zettlemoyer}]{min-etal-2020-ambigqa}
Sewon Min, Julian Michael, Hannaneh Hajishirzi, and Luke Zettlemoyer. 2020.
\newblock \href {https://doi.org/10.18653/v1/2020.emnlp-main.466} {{A}mbig{QA}: Answering ambiguous open-domain questions}.
\newblock In \emph{Proceedings of the 2020 Conference on Empirical Methods in Natural Language Processing (EMNLP)}, pages 5783--5797, Online. Association for Computational Linguistics.

\bibitem[{OpenAI(2023)}]{openai2023gpt4}
OpenAI. 2023.
\newblock \href {https://arxiv.org/abs/2303.08774} {Gpt-4 technical report}.
\newblock \emph{Preprint}, arXiv:2303.08774.

\bibitem[{Press et~al.(2023)Press, Zhang, Min, Schmidt, Smith, and Lewis}]{press-etal-2023-measuring}
Ofir Press, Muru Zhang, Sewon Min, Ludwig Schmidt, Noah Smith, and Mike Lewis. 2023.
\newblock \href {https://doi.org/10.18653/v1/2023.findings-emnlp.378} {Measuring and narrowing the compositionality gap in language models}.
\newblock In \emph{Findings of the Association for Computational Linguistics: EMNLP 2023}, pages 5687--5711, Singapore. Association for Computational Linguistics.

\bibitem[{Puerto et~al.(2024)Puerto, Tutek, Aditya, Zhu, and Gurevych}]{puertoCodePromptingElicits2024}
Haritz Puerto, Martin Tutek, Somak Aditya, Xiaodan Zhu, and Iryna Gurevych. 2024.
\newblock \href {https://arxiv.org/abs/2401.10065} {Code {{Prompting Elicits Conditional Reasoning Abilities}} in {{Text}}+{{Code LLMs}}}.
\newblock \emph{Preprint}, arxiv:2401.10065.

\bibitem[{Saeidi et~al.(2018)Saeidi, Bartolo, Lewis, Singh, Rockt{\"a}schel, Sheldon, Bouchard, and Riedel}]{saeidiInterpretationNaturalLanguage2018}
Marzieh Saeidi, Max Bartolo, Patrick Lewis, Sameer Singh, Tim Rockt{\"a}schel, Mike Sheldon, Guillaume Bouchard, and Sebastian Riedel. 2018.
\newblock \href {https://doi.org/10.18653/v1/D18-1233} {Interpretation of {{Natural Language Rules}} in {{Conversational Machine Reading}}}.
\newblock \emph{Proceedings of the 2018 Conference on Empirical Methods in Natural Language Processing}, pages 2087--2097.

\bibitem[{Servantez et~al.(2024)Servantez, Barrow, Hammond, and Jain}]{servantez2024chain}
Sergio Servantez, Joe Barrow, Kristian Hammond, and Rajiv Jain. 2024.
\newblock \href {https://arxiv.org/abs/2402.10400} {Chain of logic: Rule-based reasoning with large language models}.
\newblock \emph{Preprint}, arXiv:2402.10400.

\bibitem[{Sun et~al.(2021{\natexlab{a}})Sun, Cohen, and Salakhutdinov}]{sunConditionalQAComplexReading2021}
Haitian Sun, William~W. Cohen, and Ruslan Salakhutdinov. 2021{\natexlab{a}}.
\newblock \href {https://arxiv.org/abs/2110.06884} {{{ConditionalQA}}: {{A Complex Reading Comprehension Dataset}} with {{Conditional Answers}}}.
\newblock \emph{Preprint}, arxiv:2110.06884.

\bibitem[{Sun et~al.(2021{\natexlab{b}})Sun, Cohen, and Salakhutdinov}]{sunIterativeHierarchicalAttention2021}
Haitian Sun, William~W. Cohen, and Ruslan Salakhutdinov. 2021{\natexlab{b}}.
\newblock \href {https://doi.org/10.48550/arXiv.2106.00200} {Iterative {{Hierarchical Attention}} for {{Answering Complex Questions}} over {{Long Documents}}}.
\newblock \emph{Preprint}, arxiv:2106.00200.

\bibitem[{Sun et~al.(2022)Sun, Cohen, and Salakhutdinov}]{sun2022reasoning}
Haitian Sun, William~W. Cohen, and Ruslan Salakhutdinov. 2022.
\newblock \href {https://arxiv.org/abs/2205.12898} {Reasoning over logically interacted conditions for question answering}.
\newblock \emph{Preprint}, arXiv:2205.12898.

\bibitem[{Tao et~al.(2024)Tao, Zhao, and Feng}]{tao2024chainofdiscussion}
Mingxu Tao, Dongyan Zhao, and Yansong Feng. 2024.
\newblock \href {https://arxiv.org/abs/2402.16313} {Chain-of-discussion: A multi-model framework for complex evidence-based question answering}.
\newblock \emph{Preprint}, arXiv:2402.16313.

\bibitem[{Wang et~al.(2023)Wang, Qian, and Dou}]{wangLearningStructuredDocuments2023}
Zihan Wang, Hongjin Qian, and Zhicheng Dou. 2023.
\newblock \href {https://doi.org/10.1007/978-981-99-6207-5_3} {Learning on {{Structured Documents}} for {{Conditional Question Answering}}}.
\newblock In \emph{Chinese {{Computational Linguistics}}}, volume 14232, pages 37--57, Singapore. Springer Nature Singapore.

\bibitem[{Wei et~al.(2023)Wei, Wang, Schuurmans, Bosma, Ichter, Xia, Chi, Le, and Zhou}]{wei2023chainofthought}
Jason Wei, Xuezhi Wang, Dale Schuurmans, Maarten Bosma, Brian Ichter, Fei Xia, Ed~Chi, Quoc Le, and Denny Zhou. 2023.
\newblock \href {https://arxiv.org/abs/2201.11903} {Chain-of-thought prompting elicits reasoning in large language models}.
\newblock \emph{Preprint}, arXiv:2201.11903.

\bibitem[{Yang et~al.(2018)Yang, Qi, Zhang, Bengio, Cohen, Salakhutdinov, and Manning}]{yang2018hotpotqa}
Zhilin Yang, Peng Qi, Saizheng Zhang, Yoshua Bengio, William~W. Cohen, Ruslan Salakhutdinov, and Christopher~D. Manning. 2018.
\newblock \href {https://arxiv.org/abs/1809.09600} {Hotpotqa: A dataset for diverse, explainable multi-hop question answering}.
\newblock \emph{Preprint}, arXiv:1809.09600.

\bibitem[{Yao et~al.(2023)Yao, Zhao, Yu, Du, Shafran, Narasimhan, and Cao}]{yaoReActSynergizingReasoning2023a}
Shunyu Yao, Jeffrey Zhao, Dian Yu, Nan Du, Izhak Shafran, Karthik Narasimhan, and Yuan Cao. 2023.
\newblock \href {https://doi.org/10.48550/arXiv.2210.03629} {{{ReAct}}: {{Synergizing Reasoning}} and {{Acting}} in {{Language Models}}}.
\newblock \emph{Preprint}, arxiv:2210.03629.

\bibitem[{Ye et~al.(2023)Ye, Chen, Dillig, and Durrett}]{ye2023satlm}
Xi~Ye, Qiaochu Chen, Isil Dillig, and Greg Durrett. 2023.
\newblock \href {https://arxiv.org/abs/2305.09656} {Satlm: Satisfiability-aided language models using declarative prompting}.
\newblock \emph{Preprint}, arXiv:2305.09656.

\bibitem[{Yu et~al.(2023)Yu, Zhang, Pan, Ma, Wang, and Yu}]{yu2023chainofnoteenhancingrobustnessretrievalaugmented}
Wenhao Yu, Hongming Zhang, Xiaoman Pan, Kaixin Ma, Hongwei Wang, and Dong Yu. 2023.
\newblock \href {https://arxiv.org/abs/2311.09210} {Chain-of-note: Enhancing robustness in retrieval-augmented language models}.
\newblock \emph{Preprint}, arXiv:2311.09210.

\bibitem[{Zhang et~al.(2023)Zhang, Lin, Liu, Lai, Feng, and Zhao}]{zhang-etal-2023-many}
Chen Zhang, Jiuheng Lin, Xiao Liu, Yuxuan Lai, Yansong Feng, and Dongyan Zhao. 2023.
\newblock \href {https://doi.org/10.18653/v1/2023.findings-acl.359} {How many answers should {I} give? an empirical study of multi-answer reading comprehension}.
\newblock In \emph{Findings of the Association for Computational Linguistics: ACL 2023}, pages 5811--5827, Toronto, Canada. Association for Computational Linguistics.

\end{thebibliography}

\clearpage

\appendix
\label{sec:appendix}

\section{Datasets}
\label{app:dataset}

We use two benchmark datasets for evaluation: ConditionalQA~\cite{sunConditionalQAComplexReading2021} and ShARC~\cite{saeidiInterpretationNaturalLanguage2018}. The distribution of different question types in these datasets is presented in Table~\ref{tab:dataset_qa_types}, with additional details about the datasets provided in Table~\ref{tab:dataset_info}.

\paragraph{ConditionalQA} is a challenging benchmark for conditional question answering. It comprises a total of 3,427 questions of varying types, including yes/no questions, free-form extractive questions, questions with multiple answers, and non-answerable questions. Additionally, ConditionalQA categorizes questions into two types: \textit{deterministic}, where all necessary conditions are already satisfied within the question, and \textit{conditional}, where the complete answer must include those unsatisfied conditions as well.

\paragraph{ShARC} is a conversational QA dataset with natural language document that has conditions where questions may be underspecified, and follow-up questions are needed to finally reach the answer. And when the conditions are all satisfied, the answer could be either \textit{yes} or \textit{no}. There are some questions in it that are irrelevant to the conditions, and we discard them for simplicity. At the time we conducted our experiments, the test set was not yet publicly available, so we follow~\citet{puertoCodePromptingElicits2024} to random divide the dev set into two equal partitions and use one for experiment.

\section{LLM Setup}
\label{app:setup}

The exact models we used are as follows: GPT-3.5-0613, GPT-3.5-16k-0613, GPT-4-1106-Preview, Llama-2-70B-chat, Llama-2-13B-chat, and Mistral 7B v0.1. We ran the GPT models through the Azure AI service, and the other models on Nvidia A800. We used text-embedding-ada-002 as our retriever when comparing performance with supervised methods.

For all experiments, we used a seed of 42. The number of demonstrations for the baselines were as follows: 4 for chain of thought prompting, 3 for code prompting, and 4 for self-ask. In our approach, chain of condition, we used 4 exemplars for condition identification and logical construction, 6 for condition verification, and 4 for answer generation.

\begin{table}[t]
\centering
\small
\setlength{\tabcolsep}{5pt} 
\begin{tabular}{llc}
\toprule
\textbf{} & \textbf{Type} & \textbf{Number}\\
\midrule
\textbf{ConditionalQA} \\
\midrule
\multirow{2}{*}{Answer type} & yes/no & 1751 \\
 & extractive & 1527 \\
 \midrule
 \multirow{2}{*}{Condition type} & deterministic & 2475 \\
 & conditional & 803 \\
 \midrule
 \multirow{2}{*}{Answer number} & single & 2526 \\
 & multiple & 752 \\
 \midrule
 not answerable & & 149 \\
 \midrule
 \textbf{ShARC} \\
 \midrule
 \multirow{3}{*}{Answer type} & yes/no & 15400 \\
 & follow-up & 6814 \\
 & irrelevant & 1946 \\
\bottomrule
\end{tabular}
\caption{Question type statistics.}
\label{tab:dataset_qa_types}
\end{table}

\begin{table}[t]
\centering
\small
\setlength{\tabcolsep}{5pt} 
\begin{tabular}{lcc}
\toprule
\textbf{Dataset} & ConditionalQA & ShARC \\
\midrule
\textbf{Training} & 2338 & 21890 \\
\midrule
\textbf{Dev} & 285 & 1135 \\
\midrule
\textbf{Test} & 804 & 1135 \\
\midrule
\textbf{License} & BSD 2 & CC-BY-SA-3.0 \\
\bottomrule
\end{tabular}
\caption{Dataset details.}
\label{tab:dataset_info}
\end{table}

\section{ConditionalQA Evaluation Metrics}
\label{app:metric}

The evaluation metrics for ConditionQA include four key metrics: EM (exact match), F1, Conditional EM, and Conditional F1. EM and F1 are commonly used in QA tasks. Given a list of predicted answers \(\{\hat{a}_1,...,\hat{a}_m\}\) and a list of reference answers \(\{a_1, . . . , a_n\}\), these metrics are computed as follows:

\begin{equation}\label{EM&F1}
\footnotesize
\nonumber
\begin{array}{l}
\mathrm{EM}=\mathop{max}\limits_{\{\tilde{a}_1,...,\widetilde{a}_m\}}\frac{\sum\limits_{i=1}^{min(m,n)}s_{em}(\tilde{a}_i,a_i)\cdot\gamma_{m,n}}{n}\\ \\
\mathrm{F1}=\mathop{max}\limits_{\{\tilde{a}_1,...,\widetilde{a}_m\}}\frac{\sum\limits_{i=1}^{min(m,n)}s_{f1}(\tilde{a}_i,a_i)\cdot\gamma_{m,n}}{n}\\ \\
\gamma_{m,n}=\left\{
\begin{array}{lr}
e^{1-m/n} & \mathrm{if} \thinspace m > n \\
1 & \mathrm{if } \thinspace m\leq n\\
\end{array}\right.
\end{array}
\end{equation}

Where $\{\tilde{a}_1,...,\widetilde{a}_m\}$ is a permutation of the predicted answers $\{\hat{a}_1,...,\hat{a}_m\}$, $s_{em}$ and $s_{f1}$ are scoring functions that measures EM and F1 between two text spans. $\gamma_{m,n}$ is a penalty factor for the number of predicted answers.

While EM and F1 can evaluate the model's performance on answer prediction, they do not account for the accuracy of conditions associated with these answers. To jointly measure the performance of both answers and conditions, \citet{sunConditionalQAComplexReading2021} extended the scoring functions of EM and F1 to incorporate the prediction accuracy of conditions, resulting in two new metrics: Conditional EM and Conditional F1. These new scoring functions are computed as follows:

\begin{equation}\label{CondEM&F1}
\footnotesize
\nonumber
\begin{array}{l}
s_{em+c}(\tilde{a}_i,\tilde{C}_i,a_i,C_i)=s_{em}(\tilde{a}_i,a_i)\cdot \mathrm{F1}(\tilde{C}_i,C_i) \\ \\
s_{f1+c}(\tilde{a}_i,\tilde{C}_i,a_i,C_i)=s_{f1}(\tilde{a}_i,a_i)\cdot \mathrm{F1}(\tilde{C}_i,C_i) \\ \\
\mathrm{EM_{+c}}=\mathop{max}\limits_{\{\tilde{a}_1,...,\widetilde{a}_m\}}\frac{\sum\limits_{i=1}^{min(m,n)}s_{em+c}(\tilde{a}_i,\tilde{C}_i,a_i,C_i)\cdot\gamma_{m,n}}{n}\\ \\
\mathrm{F1_{+c}}=\mathop{max}\limits_{\{\tilde{a}_1,...,\widetilde{a}_m\}}\frac{\sum\limits_{i=1}^{min(m,n)}s_{f1+c}(\tilde{a}_i,\tilde{C}_i,a_i,C_i)\cdot\gamma_{m,n}}{n}\\ \\
\end{array}
\end{equation}

Here, \(\tilde{C}_i\) represents the set of conditions predicted by the model corresponding to the answer \(\tilde{a}_i\), and \(C_i\) represents the oracle (ground truth) set of conditions. \(\mathrm{F1}(\tilde{C}_i, C_i)\) denotes the HTML element level F1 score between the predicted set of conditions and the oracle set of conditions.

\begin{table}[t]
\centering
\small
\begin{tabular}{l|cccc}
\toprule
 & \textbf{FP} & \textbf{IE} & \textbf{VE} & \textbf{Total} \\
\midrule
\#Conds & 57.8 & 55.0 & 17.5 & 130.3 \\
Ratio & 44.4\% & 42.2\% & 13.4\% & 100\% \\
\bottomrule
\end{tabular}
\caption{Prediction error on ConditionalQA.}
\label{tab:error_conditionalqa}
\end{table}

\begin{table}[t]
\centering
\small
\begin{tabular}{l|cccc}
\toprule
\textbf{Model} & \textbf{GPT} & \textbf{Llama(70b)} & \textbf{Llama(13b)} & \textbf{Mistral} \\
\midrule
\textbf{Recall} & 67\% & 63\% & 52\% & 39\% \\
\bottomrule
\end{tabular}
\caption{Performance of condition identification.}
\label{tab:error_identification}
\end{table}

\begin{table}[t]
\centering
\small
\begin{tabular}{l|ccc}
\toprule
 & \textbf{Step 1} & \textbf{Step 2} & \textbf{Step 3} \\
\midrule
GPT-3.5 (api) & 2.5 & 2.8 & 1.8 \\
Llama-2-70b (vllm) & 29.9 & 63.2 & 28.7 \\
Llama-2-13b (vllm) & 6.0 & 6.1 & 3.8 \\
Mistral & 27.2 & 78.1 & 28.7 \\
\bottomrule
\end{tabular}
\caption{Run-time efficiency of our method. Where \textit{step 1} stands for Condition Identification and Logical Construction, \textit{step 2} for Condition Verification, and \textit{step 3} for Expression Solution and Answer Generation.}
\label{tab:runtime}
\end{table}

\section{Error Analysis}
\label{app:error}

We investigate the prediction errors in ConditionalQA. We report detailed statistics for condition prediction. Errors are classified into False Positive (FP) and False Negative (FN) categories. Since chain of condition explicitly identifies all conditions, False Negatives can be further classified into Identifying Errors (IE) and Verification Errors (VE) based on the step at which the model makes mistakes. The results, averaged across four models, are shown in Table~\ref{tab:error_conditionalqa}.

We found that most errors occur during the condition identification step. Consequently, we further investigate its performance, as shown in Table~\ref{tab:error_identification}. The results indicate that the performance of the condition identification step remains unsatisfactory.

\section{Run-Time Efficiency}
\label{app:runtime}

We report the run-time efficiency for GPT-3.5, Llama-2-70b, Llama-2-13b, and Mistral-7b averaged on two datasets in Table~\ref{tab:runtime}. We calculated the second per iteration (s/it) as a metric for runtime efficiency.

\section{More Detailed Results}
\label{app:results}

We report the detailed results on ConditionalQA according to different question types in Table~\ref{tab:detailed_result_GPT} for GPT-3.5, Table~\ref{tab:detailed_result_llama70b} for Llama-2 (70B), Table~\ref{tab:detailed_result_llama13b} for Llama-2 (13B) and Table~\ref{tab:detailed_result_mistral} for Mistral.

\section{Prompt Examples}
\label{app:prompt}

We provide an example of the prompt for condition identification and logical construction in Table~\ref{tab:conditionalqa_prompt_step1} for ConditionalQA and in Table~\ref{tab:sharc_prompt_step1} for ShARC. We provide an example for condition verification in Table~\ref{tab:conditionalqa_prompt_step2} for ConditionalQA and in Table~\ref{tab:sharc_prompt_step2} for ShARC. We provide an example for answer generation in Table~\ref{tab:conditionalqa_prompt_step3} for ConditionalQA and in Table~\ref{tab:sharc_prompt_step3} for ShARC.

\begin{table*}[!t]
\footnotesize
\centering
\setlength\tabcolsep{4pt}
\begin{tabular}{lcccccccc}
\toprule
\multirow{2}{*}{} & \multicolumn{2}{c}{\textbf{Yes/No}} & \multicolumn{2}{c}{\textbf{Extractive}} & \multicolumn{2}{c}{\textbf{Conditional}} & \multicolumn{2}{c}{\textbf{Overall}} \\
 & \textbf{EM/F1} & \textbf{w/conds} & \textbf{EM/F1} & \textbf{w/conds} & \textbf{EM/F1} & \textbf{w/conds} & \textbf{EM/F1} & \textbf{w/conds} \\
\midrule
Zero-Shot & 82.1/82.1 & 17.0/17.0 & 29.5/55.0 & 19.7/32.1 & 40.7/49.1 & 12.9/16.0 & 59.5/71.0 & 23.9/29.5 \\
Chain of Thought & 80.4/80.4 & 54.2/54.2 & 31.3/55.0 & 29.6/\textbf{50.1} & 45.8/53.6 & 13.1/16.4 & 59.3/70.0 & 45.4/54.6 \\
Code Prompting & 81.1/81.1 & 62.6/62.6 & 32.9/50.2 & \textbf{32.2}/47.2 & 47.2/54.3 & 8.5/11.5 & 60.4/68.2 & 50.8/57.5  \\
Self-Ask & 76.2/76.2 & 49.8/49.8 & 26.1/52.8 & 25.3/49.7 & 49.7/58.3 & 13.5/17.5 & 54.9/66.9 & 41.3/52.2 \\
Chain of Condition & \textbf{87.4}/\textbf{87.4} & \textbf{67.1}/\textbf{67.1} & \textbf{35.2}/\textbf{55.6} & 31.8/50.0 & \textbf{56.0}/\textbf{62.2} & \textbf{18.9}/\textbf{20.7} & \textbf{64.6}/\textbf{73.7} & \textbf{52.9}/\textbf{61.0} \\
\bottomrule
\end{tabular}
\caption{Result of different prompting methods on GPT-3.5-Turbo.}
\label{tab:detailed_result_GPT}
\end{table*}

\begin{table*}[!t]
\footnotesize
\centering
\setlength\tabcolsep{4pt}
\begin{tabular}{lcccccccc}
\toprule
\multirow{2}{*}{} & \multicolumn{2}{c}{\textbf{Yes/No}} & \multicolumn{2}{c}{\textbf{Extractive}} & \multicolumn{2}{c}{\textbf{Conditional}} & \multicolumn{2}{c}{\textbf{Overall}} \\
 & \textbf{EM/F1} & \textbf{w/conds} & \textbf{EM/F1} & \textbf{w/conds} & \textbf{EM/F1} & \textbf{w/conds} & \textbf{EM/F1} & \textbf{w/conds} \\
\midrule
Zero-Shot & 68.2/68.2 & 36.9/36.9 & 10.8/27.0 & 7.1/16.7 & 52.9/55.7 & 9.2/9.5 & 44.0/51.2 & 26.6/30.9 \\
Chain of Thought & 78.2/78.2 & 49.1/49.1 & \textbf{40.1}/\textbf{60.6} & \textbf{35.5}/\textbf{53.7} & 48.7/53.8 & 11.4/13.6 & 62.2/71.4 & 45.5/53.7 \\
Code Prompting & 76.2/76.2 & 17.0/17.0 & 24.9/44.3 & 5.4/12.9 & 56.9/61.1 & \textbf{19.6}/\textbf{21.2} & 54.4/63.1 & 15.9/19.2  \\
Self-Ask & 79.7/79.7 & 35.4/35.4 & 31.8/55.6 & 30.0/50.9 & 53.1/61.6 & 17.6/20.9 & 59.2/69.9 & 36.1/45.5 \\
Chain of Condition & \textbf{84.5}/\textbf{84.5} & \textbf{54.8}/\textbf{54.8} & 35.4/60.4 & 32.2/52.0 & \textbf{49.3}/\textbf{57.0} & 17.4/19.6 & \textbf{64.7}/\textbf{75.2} & \textbf{47.7}/\textbf{56.0} \\
\bottomrule
\end{tabular}
\caption{Result of different prompting methods on Llama-2(70B).}
\label{tab:detailed_result_llama70b}
\end{table*}

\begin{table*}[!t]
\footnotesize
\centering
\setlength\tabcolsep{4pt}
\begin{tabular}{lcccccccc}
\toprule
\multirow{2}{*}{} & \multicolumn{2}{c}{\textbf{Yes/No}} & \multicolumn{2}{c}{\textbf{Extractive}} & \multicolumn{2}{c}{\textbf{Conditional}} & \multicolumn{2}{c}{\textbf{Overall}} \\
 & \textbf{EM/F1} & \textbf{w/conds} & \textbf{EM/F1} & \textbf{w/conds} & \textbf{EM/F1} & \textbf{w/conds} & \textbf{EM/F1} & \textbf{w/conds} \\
\midrule
Zero-Shot & 66.4/66.4 & 35.7/35.7 & 9.1/25.3 & 6.3/14.4 & \textbf{51.0}/\textbf{54.1} & 6.2/7.2 & 42.3/49.6 & 26.1/28.9 \\
Chain of Thought & 69.7/69.7 & 40.5/40.5 & \textbf{37.5}/\textbf{57.5} & \textbf{29.9}/43.6 & 42.0/50.2 & 10.6/13.5 & 56.8/65.8 & 38.7/44.8 \\
Code Prompting &  65.7/65.7 & 8.5/8.5 & 17.7/26.3 & 4.1/6.9 & 49.5/51.9 & 11.8/12.1 & 45.9/49.7 & 11.0/12.3 \\
Self-Ask & 65.7/65.7 & 34.4/34.4 & 22.2/48.9 & 18.1/36.0 & 35.2/41.5 & 5.7/7.5 & 47.9/59.9 & 30.3/38.3 \\
Chain of Condition & \textbf{77.6}/\textbf{77.6} & \textbf{52.5}/\textbf{52.5} & 29.8/51.6 & 26.2/\textbf{44.8} & 45.8/52.6 & \textbf{13.8}/\textbf{15.4} & \textbf{57.2}/\textbf{67.1} & \textbf{43.0}/\textbf{51.3}\\
\bottomrule
\end{tabular}
\caption{Result of different prompting methods on Llama-2(13B).}
\label{tab:detailed_result_llama13b}
\end{table*}

\begin{table*}[!t]
\footnotesize
\centering
\setlength\tabcolsep{4pt}
\begin{tabular}{lcccccccc}
\toprule
\multirow{2}{*}{} & \multicolumn{2}{c}{\textbf{Yes/No}} & \multicolumn{2}{c}{\textbf{Extractive}} & \multicolumn{2}{c}{\textbf{Conditional}} & \multicolumn{2}{c}{\textbf{Overall}} \\
 & \textbf{EM/F1} & \textbf{w/conds} & \textbf{EM/F1} & \textbf{w/conds} & \textbf{EM/F1} & \textbf{w/conds} & \textbf{EM/F1} & \textbf{w/conds} \\
\midrule
Zero-Shot & 68.5/68.5 & 36.3/36.3 & 10.8/25.4 & 7.9/17.0 & \textbf{55.3}/56.9 & 4.7/5.1 & 44.1/50.7 & 26.7/30.8\\
Chain of Thought & \textbf{80.5}/\textbf{80.5} & 41.2/41.2 & \textbf{28.9}/\textbf{51.8} & \textbf{26.9}/\textbf{46.3} & 51.4/\textbf{59.9} & 15.9/\textbf{18.7} & \textbf{58.3}/\textbf{68.6} & 37.7/46.4 \\
Code Prompting & 72.7/72.7 & 10.1/10.1 & 15.5/24.3 & 1.0/1.5 & 53.9/55.9 & \textbf{16.9}/17.3 & 48.4/52.3 & 10.4/10.6 \\
Self-Ask & 74.1/74.1 & \textbf{58.3}/\textbf{58.3} & 16.7/40.8 & 15.6/36.2 & 41.9/50.7 & 13.7/17.0 & 49.6/60.5 & \textbf{41.2}/\textbf{50.4} \\
Chain of Condition & 80.4/80.4 & 51.9/51.9 & 22.8/41.3 & 21.7/36.8 & 43.4/48.1 & 9.0/11.1 & 55.5/63.8 & 40.7/47.5 \\
\bottomrule
\end{tabular}
\caption{Result of different prompting methods on Mistral.}
\label{tab:detailed_result_mistral}
\end{table*}

\begin{table*}
\centering
\small
\setlength{\tabcolsep}{5pt}
\begin{tabular}{p{2.0\columnwidth}r}
\toprule
- Please find all the paragraphs that contain conditions for the suggested answer based on the provided documents.\\
- Provided Documents:\\
<h1>Joining a workplace pension</h1>\\
<p>All employers must provide a workplace pension scheme. This is called ‘automatic enrolment’.</p>\\
<p>Your employer must automatically enrol you into a pension scheme and make contributions to your pension if all of the following apply:</p>\\
<li>you’re classed as a ‘worker’</li>\\
<li>you’re aged between 22 and State Pension age</li>\\
<li>you earn at least £10,000 per year</li>\\
<li>you usually (‘ordinarily’) work in the UK (read the detailed guidance if you’re not sure)</li>\\
- Question: I am 23 years old. I have just got a new job in UK that will have a salary of £18,000 per year. I do not want to pay into a pension at this time. Does my employer need to automatically enrol me onto a pension?\\
- Suggested Answer: yes\\
- Response: \\
Reasoning: The user is asking whether the employer needs to automatically enrol the user onto a pension. According to the documents, the employer need automatically enrol the user into a pension scheme if the user is classed as a worker, aged between 22 and State Pension age, earns at least £10,000 per year, and usually works in the UK. So these are the prerequisites that all need to be satisfied, they are placed next to each other in the documents and are all needed to be satisfied, so they are in the same group and the relationship between them is "and". \\
Conditions: [\{"contents": ["<li>you’re classed as a ‘worker’</li>", "<li>you’re aged between 22 and State Pension age</li>", "<li>you earn at least £10,000 per year</li>", "<li>you usually (‘ordinarily’) work in the UK (read the detailed guidance if you’re not sure)</li>"], "relationship": "and"\}] \\
\bottomrule
\end{tabular}
\caption{ConditionalQA prompt example for condition identification and logical construction.}
\label{tab:conditionalqa_prompt_step1}
\end{table*}

\begin{table*}
\centering
\small
\setlength{\tabcolsep}{5pt}
\begin{tabular}{p{2.0\columnwidth}r}
\toprule
- Please check the relevance, find the condition and check the satisfaction of the condition.\\
- Reference Document:\\
<h1>Who must send a tax return</h1>\\
<p>You must send a tax return if, in the last tax year (6 April to 5 April), you were:</p>\\
<li>self-employed as a 'sole trader' and earned more than \$31,000  (before taking off anything you can claim tax relief on)</li>\\
<li>a partner in a business partnership</li>\\
<p>You will not usually need to send a return if your only income is from your wages or pension. But you may need to send one if you have any other untaxed income, such as:</p>\\
<li>money from renting out a property</li>\\
<li>tips and commission</li>\\
- Segment to Check: <li>self-employed as a 'sole trader' and earned more than \$31,000  (before taking off anything you can claim tax relief on)</li>\\
- Question: I decided to start freelancing and working for myself and made freelancing my main job, taxes are not automatically deducted from my income. I have an income from my job for this tax year. How do I pay taxes on my income?\\
- Suggested Answer: send a tax return\\
- Response: \\
Reasoning: The user is asking how to pay taxes on his income. The segment is about a prerequisite, and by looking at the reference document, we know this prerequisite is for the user to pay taxes return, so this segment is relevant to the question and answer, and contains a condition that the user needs to be self-employed as a "sole trader" and earned more than \$31,000. Then let's check the satisfaction of this in the question. We can see in the question, the user mentions that he decided to start freelancing and working for himself and made freelancing his main job, and he has an income for this tax year. But he does not explicitly mention the amount of income he earned, so we could not determine whether he earned more than \$31,000, and the satisfaction of this condition is "not mentioned".\\
Satisfaction: not mentioned\\
\bottomrule
\end{tabular}
\caption{ConditionalQA prompt example for condition verification.}
\label{tab:conditionalqa_prompt_step2}
\end{table*}

\begin{table*}
\centering
\small
\setlength{\tabcolsep}{5pt}
\begin{tabular}{p{2.0\columnwidth}r}
\toprule
- Please answer the question based on the provided documents and known information.\\
- Provided Documents:\\
<h1>Joining a workplace pension</h1>\\
<p>All employers must provide a workplace pension scheme. This is called ‘automatic enrolment’.</p>\\
<p>Your employer must automatically enrol you into a pension scheme and make contributions to your pension if all of the following apply:</p>\\
<li>you’re aged between 22 and State Pension age</li>\\
<li>you earn at least £10,000 per year</li>\\
<li>you usually (‘ordinarily’) work in the UK (read the detailed guidance if you’re not sure)</li>\\
- Question: I am 23 years old. I have just got a new job in UK that will have a salary of £8,000 per year. I do not want to pay into a pension at this time. Does my employer must enrol me onto a pension?\\
- Known Information: Q: You're aged between 22 and State Pension age? A: Yes; Q: You earn at least £10,000 per year? A: No; Q: You usually (‘ordinarily’) work in the UK (read the detailed guidance if you’re not sure)? A: Yes\\
- Answer Type: yes/no\\
- Response: \\
Reasoning: The user is asking whether the employer needs to automatically enrol the user onto a pension. In the documents, the employer must automatically enrol the user into a pension scheme if the user is aged between 22 and State Pension age, and earns at least £10,000 per year, and usually works in the UK. In the known information, we know the user is aged between 22 and State Pension age, and usually works in the UK, but the user earns less than £10,000 per year. Therefore, the user does not satisfy the requirement of earning at least £10,000 per year, so the answer to the question is "no".\\
Answer: No\\
\bottomrule
\end{tabular}
\caption{ConditionalQA prompt example for answer generation.}
\label{tab:conditionalqa_prompt_step3}
\end{table*}

\begin{table*}
\centering
\small
\setlength{\tabcolsep}{5pt}
\begin{tabular}{p{2.0\columnwidth}r}
\toprule
- Please find all conditions based on the provided documents.\\
- Provided Documents:\\
6. Immigration control\\
You may be able to get Child Tax Credit if you’re either working in the UK or you’re not working because you’re:\\
* retired\\
* pregnant or looking after children\\
* sick or disabled or your partner has died\\
- Question: I was born in North Carolina and my husband and I are still trying to conceive. May I be able to get Child Tax Credit?\\
- Response: \\
Reasoning: The user is asking whether she may be able to get Child Tax Credit. According to the documents, you may be able to get Child Tax Credit if you’re either working in the UK, or if you are not working because you’re retired, pregnant or looking after children, sick or disabled, or your partner has died. So there are precondition that the user needs to satisfy before she may be able to get Child Tax Credit, and there relationship is "or".\\
Conditions: \{"contents": ["You're working in the UK", "You're not working because you're retired", "You're not working because you're pregnant", "You're not working because you're looking after children", "You're not working because you're sick", "You're not working because you're disabled", "Your partner has died"], "relationship": "or"\}\\
\bottomrule
\end{tabular}
\caption{ShARC prompt example for condition identification and logical construction.}
\label{tab:sharc_prompt_step1}
\end{table*}

\begin{table*}
\centering
\small
\setlength{\tabcolsep}{5pt}
\begin{tabular}{p{2.0\columnwidth}r}
\toprule
- Please check the satisfaction of the given condition("satified" or "contradicted" or "not mentioned") based on the provided documents.\\
- Provided Documents:\\
6. Immigration control\\
You may be able to get Child Tax Credit if you’re either working in the UK or you’re not working because you’re:\\
* retired\\
* pregnant or looking after children\\
* sick or disabled or your partner has died\\
- Condition to Check: You\'re working in the UK\\
- Known Information: Q: Are you working in the UK? A: No\\
- Question: I was born in North Carolina and my husband and I are still trying to conceive. May I be able to get Child Tax Credit?\\
- Response: \\
Reasoning: The condition to check is whether the user is working in the UK. According to the known information about the user, she mentioned that she is not working in the UK. So the condition is contradicted by the known information.\\
Satisfaction: contradicted\\
\bottomrule
\end{tabular}
\caption{ShARC prompt example for condition verification.}
\label{tab:sharc_prompt_step2}
\end{table*}

\begin{table*}
\centering
\small
\setlength{\tabcolsep}{5pt}
\begin{tabular}{p{2.0\columnwidth}r}
\toprule
- Please answer the question based on the provided document.\\
- Provided Document:\\
6. Immigration control\\
You may be able to get Child Tax Credit if you’re either working in the UK or you’re not working because you’re:\\
* retired\\
* pregnant or looking after children\\
* sick or disabled or your partner has died\\
- Known Information: Q: Are you working in the UK? A: No; Q: You are retired? A: No; Q: You are pregnant? A: No; Q: You are looking after children? A: No; Q: You are sick or disabled? A: Yes\\
- Question: I was born in North Carolina and my husband and I are still trying to conceive. May I be able to get Child Tax Credit?\\
- Response: \\
Reasoning: The user is asking whether she may be able to get Child Tax Credit. According to the documents, you may be able to get Child Tax Credit if you’re either working in the UK, or if you are not working because you’re retired, pregnant or looking after children, sick or disabled, or your partner has died. In the known information, the user mentioned that she is not working in the UK, she is not retired or pregnant or looking after children, but she is sick or disabled. So she is still eligible to get Child Tax Credit.\\
Answer: Yes\\
\bottomrule
\end{tabular}
\caption{ShARC prompt example for answer generation.}
\label{tab:sharc_prompt_step3}
\end{table*}

\end{document}